\documentclass[a4paper,12pt]{article}

\usepackage[dvipsnames]{xcolor}
\usepackage{amsmath}
\usepackage{amssymb}
\usepackage{dsfont}
\usepackage[english]{babel}
\usepackage[utf8]{inputenc}
\usepackage{times}
\usepackage{graphics}
\usepackage{graphicx}
\usepackage[T1]{fontenc}
\usepackage[official,right]{eurosym}
\usepackage{url}
\usepackage{eso-pic}
\usepackage{tikz}
\usetikzlibrary {arrows.meta,bending,positioning}

\usepackage{eurosym}
\usepackage{enumerate,rotating,multirow,xcolor,booktabs,placeins}
\usepackage{algorithm}

\usepackage{algpseudocode}

\usepackage{enumitem}
\newlist{steps}{enumerate}{1}
\setlist[steps, 1]{label = Step \arabic*:}

\newtheorem{corollary}{Corollary}[section]

\newtheorem{theorem}{Theorem}[section]

\newtheorem{example}{Example}[section]
\def\qed{\hbox to\hsize{\hfill\vrule height 1.6ex width 1.5ex depth -.1ex}}

\newtheorem{definition}{Definition}[section]
\newtheorem{proposition}{Proposition}[section]

\title{Conflict Transformation and management. \\ From Cognitive Maps to Value Trees}
\author{Berkay H. Tosunlu$^\ddag$, Joseph H.A. Guillaume$^\dag$, Alexis Tsouki\`as$^\ddag$ \\ $^\dag$Fenner School of Environment, Australian National University \\ $^\ddag$CNRS-LAMSADE, PSL, Universit\'e  Paris Dauphine}
\date{}

\begin{document}

\thispagestyle{empty}

\enlargethispage*{8cm}
 \vspace*{-38mm}

\AddToShipoutPictureBG*{\includegraphics[width=\paperwidth,height=\paperheight]{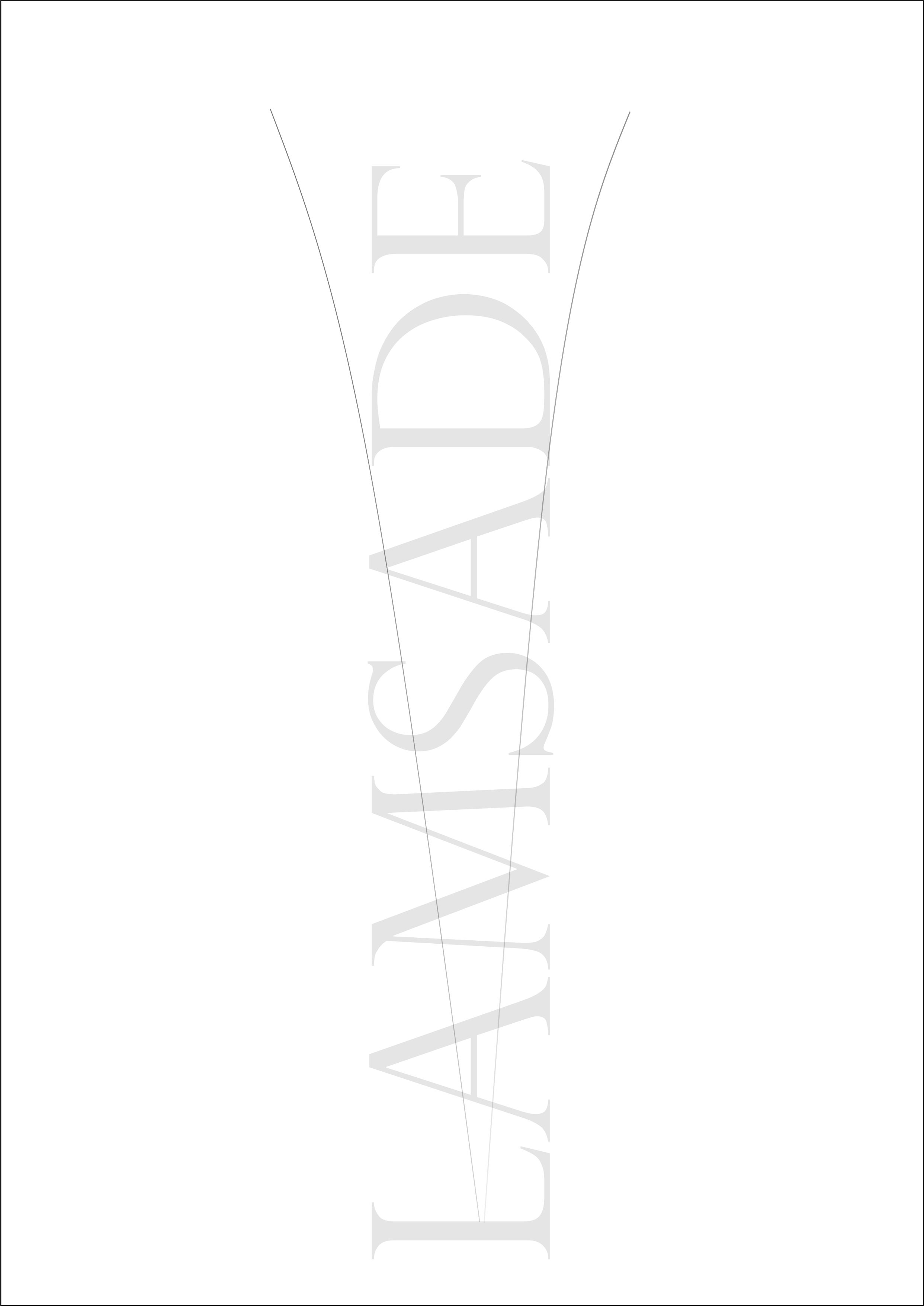}}

\begin{minipage}{24cm}
 \hspace*{-28mm}
\begin{picture}(500,700)\thicklines
 \put(60,670){\makebox(0,0){\scalebox{0.7}{\includegraphics{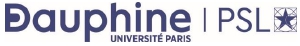}}}}
 \put(60,70){\makebox(0,0){\scalebox{0.3}{\includegraphics{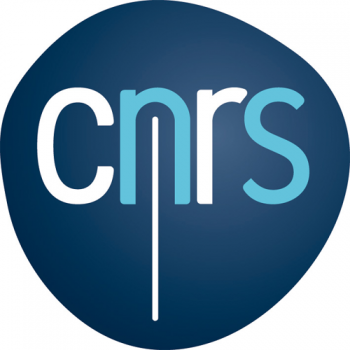}}}}
 \put(320,350){\makebox(0,0){\Huge{CAHIER DU \textcolor{BurntOrange}{LAMSADE}}}}
 \put(140,10){\textcolor{BurntOrange}{\line(0,1){680}}}
 \put(190,330){\line(1,0){263}}
 \put(320,310){\makebox(0,0){\Huge{\emph{405}}}}
 \put(320,290){\makebox(0,0){December 2023}}
 \put(320,210){\makebox(0,0){\Large{Conflict Transformation and management.}}}
 \put(320,190){\makebox(0,0){\Large{From Cognitive Maps to Value Trees}}}
 \put(320,100){\makebox(0,0){\Large{Berkay H. Tosunlu, Joseph H.A. Guillaume,}}}
 \put(320,80){\makebox(0,0){\Large{Alexis Tsouki\`as}}}
 \put(320,670){\makebox(0,0){\Large{\emph{Laboratoire d'Analyse et Mod\'elisation}}}}
 \put(320,650){\makebox(0,0){\Large{\emph{de Syst\`emes pour l'Aide \`a la D\'ecision}}}}
 \put(320,630){\makebox(0,0){\Large{\emph{UMR 7243}}}}
\end{picture}
\end{minipage}

\newpage

\addtocounter{page}{-1}

\maketitle


\begin{abstract}

Conflict transformation and management are complex decision processes with extremely high stakes at hand and could greatly benefit from formal approaches to decision support. For this purpose we develop a general framework about how to use problem structuring methods for such purposes. More precisely we show how to transform cognitive maps to value trees in order to promote a more design-oriented approach to decision support aiming at constructing innovative solutions for conflict management purposes. We show that our findings have a much wider validity since they allow to move from a descriptive representation of a problem situation to a more prescriptive one using formal procedures and models.

\end{abstract}

%
%
%
\newpage

\section{Introduction}

Conflicts are natural. It is natural that people, organisations, institutions, classes, groups, states, have differences about diverse stakes at diverse moments. By itself, conflict is not a problem and often can be an opportunity for creativity and innovation. What is a problem, is the violent degeneration of conflicts, their destructive issue, where people, goods, infrastructures, get lost, most of the times without even ending the conflict. This simple observation, together with the increase and diversification of conflicts, led to establishing an interdisciplinary field of research about ``Conflict Transformation and Peace Studies'' (see \cite{Galtung1969}, \cite{Galtung1976}, \cite{Korppenetal2011}, \cite{RogersRamsbotham1999}). We are not going to survey this extensive literature here (it is not the scope of this paper), but want to mention that Decision Analysts and Operational Researchers contributed and contribute to this area (see \cite{BartolucciGallo2010}, \cite{Gallo2013}, \cite{Hipeletal2020}).

Under such a perspective our paper is a contribution of Decision Analysis to the study of Conflict Transformation and Management. However, our contribution can be considered as broader than this ``application domain'', since the theoretical part is more general. More precisely, we explore two problem structuring methods, disconnected until today, as combined tools for the purpose of supporting conflict transformation and management processes aiming at establishing the possibility of constructive issue of a conflict.

Problem Solving Methods (PSMs) have been designed as a branch of Decision Analysis and Operational Research (for some as an alternative), claiming that understanding a problem is as important as solving it (see \cite{Ackoff79a}, \cite{Ackoff79b}, \cite{check1book81}, \cite{FrancoShawWestcombe2006}, \cite{rosenheadmingers2001}, \cite{ShawFrancoWestcombe2007}). In this paper we focus on two well known such PSMs: Cognitive Mapping (see \cite{EdenJonesSims83}) and Value Trees (see \cite{Keeney92}) which we consider relevant for the purpose of supporting conflict transformation and management processes. We need however, to make explicit a critical perspective about these two methods.

Most PSMs are essentially descriptive. They help in understanding who has which problem and why. Therefore PSMs are mostly designed in order to assist different stakeholders, involved in complex problem situations, by developing comprehensive assessments and establishing common ground between the parties (see \cite{Ackermann2012}). This is certainly extremely useful, but is not prescriptive: it does not help in suggesting how to ``practically'' work out a ``solution'' (if any) for the problem situation where the stakeholders are involved.

Notable exceptions are the ``Strategic Choice Approach'' (see \cite{FriendHickling87}) and the ``Value Trees'' (see \cite{Keeney92}). However, such methods do little in order to understand how the problem situation is structured and how decision aiding can be conducted (see \cite{Tsoukias07aor}, \cite{Tsoukias08ejor}). More recently there have been contributions emphasising the necessity to be more ``design oriented'' in conducting decision support activities (see \cite{ColorniTsoukias2020}, \cite{FerrettietalEJOR2018}, \cite{PluchinottaetalGDN2019}). However, we still need more operational directions on how we can move from describing a problem situation to how we can design alternative solutions; what to do after we understood what the problem is. In this paper we argue that PSMs should not only be descriptive but also design-oriented.

Drawing from Kelly's theory (\cite{kelly1955psychology}, of human beings as problem solvers, the definition of cognitive mapping given by Eden (see \cite{eden1988cognitive}) emphasizes the subjective nature of problem handling, where the problem owner's personal understanding and definition of the problem are considered essential. This approach acknowledges that the problem owner's perception of why a situation is problematic, and how it relates to their system of values, plays a significant role in problem-solving, even if the owner may not have a ready solution in mind. Cognitive mapping focuses on describing the problem owner's cognitive system, including their values, beliefs, and objectives, rather than simply analyzing external variables (\cite{eden1994cognitive}, \cite{kelly1955psychology}). While cognitive mapping provides a suitable means of problem description, it may not inherently provide specific strategies for resolving or managing the conflict, necessitating additional steps in the problem-solving process.

On the other hand, value trees are a way to structure problems in a formal manner and reveal values in a more structured and hierarchical way. Keeney's value-focused thinking suggests that decision-making should focus on values rather than just solving problems considering alternatives as given (\cite{keeney1994creativity}). By considering values, new and innovative solutions can emerge. Value trees are a common formal technique used to represent value-focused thinking and reveal values in a structured way. Despite providing a means for generating new alternatives, value trees may not always fully capture the complexity of the problem structure due to the absence of a universally agreed-upon method for constructing them. Consequently, the use of value trees may lead to information loss and attribute asymmetry (see \cite{jacobi2007quantifying}, \cite{poyhonen1998notes}), which should be carefully considered in any decision-making process.

Our main objective is to combine cognitive maps and value trees and to develop them into effective policy design tools for conflict transformation and management. This involves two main objectives. First, we aim to improve the structures of these two well-known Problem Structuring Methods integrating them in a synergistic manner (as suggested in  \cite{marttunen2017structuring}). By combining the strengths of both methods, we aim to create a more robust and more comprehensive approach to problem structuring that takes into account both the subjective understanding and values of the problem owner (as emphasized in cognitive mapping) and the formalized value-oriented thinking (as advocated in value trees). Second, we aim to extract a value tree from the cognitive map, integrating the insights gained from the cognitive mapping process into the formalized structure of a value tree. This would enable us to leverage the rich qualitative information captured in cognitive maps and translate it into a more structured and formal representation of values that can be used for policy design and decision-making.

This paper delves into the intricate relationship between cognitive maps and value trees, with a focus on how the graph structure of cognitive maps can be transformed into the tree structure of values. We introduce the concept of a ``value cognitive map'' as an intermediate step that sheds light on the simplifications and assumptions involved in the conversion from a cognitive map to a value tree. By retaining the descriptive structure of the cognitive map while incorporating the insights from value-oriented thinking and value trees, our proposed approach not only aids in conflict resolution but also serves as a valuable tool for decision-making. The integration of cognitive maps and value trees expands the set of alternative solutions by incorporating new perspectives and considerations from both the problem owner and the value-oriented thinking process. This makes our proposed structure relevant not only in conflicting situations but more generally in decision-aiding, as it enables a more comprehensive analysis of the problem space while preserving the problem owner's perspective. By combining the strengths of cognitive maps and value trees, our approach offers a promising framework for enhancing problem-structuring methods and advancing policy design and decision-making processes.

In our study, we applied our proposed approach of combining cognitive maps and value trees to the context of the Kurdish-Turkish conflict. Through our analysis, we found that employing specific approaches in building the cognitive map can greatly facilitate the transformation process and aid in identifying common ground and potential areas for compromise. By integrating the insights gained from the cognitive map and the value-oriented thinking process inherent in value trees, we were able to generate new alternatives for each perspective that could potentially contribute to a conflict resolution. Our findings highlight the effectiveness of our approach in not only providing a comprehensive problem analysis, but also in fostering a collaborative mindset by uncovering shared values and potential areas of agreement. This underscores the potential of our approach as a valuable tool in policy design and conflict management efforts, particularly in complex and contentious situations such as the Kurdish-Turkish conflict.

The structure of this paper is the following: Section 2 covers the literature on cognitive mapping and value trees, along with the background of our motivation. In Section 3, we present our proposal, detailing the transformation process from a cognitive map to a value cognitive map, then from a value cognitive map to an ends-means map, and then from an ends-mean map to a value tree. In Section 4, we apply this method to analyze the Kurdish-Turkish conflict. Conclusions summarise our findings and present further research directions.

\section{Literature and background}

\subsection{Cognitive mapping}

The concept of a cognitive map has been defined in various ways in the existing literature, reflecting its interdisciplinary nature and wide-ranging applications. First introduced as a term by Tolman in 1948 (\cite{tolman1948cognitive}), cognitive maps have since been extensively utilized across diverse disciplines, including psychology and natural sciences [29], management (see \cite{ackermann2001soda}, \cite{andersen1997international}, \cite{cossette1994cartes}, \cite{fiol1992maps}, \cite{ferreira2016experience}, \cite{langfield1992exploring}, \cite{lee1992system}, \cite{robertson2006understanding}, \cite{weick1986organizations}), politics (see \cite{arvidson1995cognitive}, \cite{axelrod2015structure}, \cite{eden2004cognitive}, \cite{kwahk1999supporting}), economics (see \cite{ccoban2005prediction}, \cite{gupta2017modeling}, \cite{kauko2002modelling}, \cite{oliveira2017integrating}, \cite{papageorgiou2019fuzzy}) and other areas (see \cite{arvidson1995cognitive}, \cite{axelrod2015structure}). In the problem structuring literature, cognitive mapping has been employed in different contexts, as evidenced in several papers (see \cite{borroi1998relationship}, \cite{carlsson1997cognitive}, \cite{ccoban2005prediction}, \cite{damart2010cognitive}, \cite{eden2004analyzing}, \cite{eden2004cognitive}, \cite{kwahk1999supporting}, \cite{lee1992system}, \cite{nakagawa2010assessment}, \cite{csahin2004using}, \cite{swan1997using}, \cite{tegarden2003group}, \cite{verstraete1996cartographie} among others). While cognitive mapping is widely recognized as a problem-oriented approach to decision support, with structures on the map expected to be propositional and indicative of actions, it is noteworthy that the definitions of cognitive maps in the literature are primarily functional in nature, focusing on their role as a problem structuring method, rather than being design-oriented. Notably, the definition put forth by Eden, inspired by Kelly's ``theory of human beings as problem solvers'', is widely accepted as a descriptive tool to elucidate how problem owners perceive and describe the problem at hand.

A more critical perspective of the theory and practice of using cognitive maps reveals a number of limitations at least as far as their use for conflict transformation and management purposes is concerned.


\begin{enumerate}

\item Firstly, one limitation of cognitive mapping in conflict transformation and management is that it primarily provides a descriptive understanding of the problem owner's perception of the problem. It focuses on how the problem owner structures and organizes information about the problem, but it may not necessarily offer innovative or creative solutions for resolving the conflict. While understanding the problem owner's perspective is important, it may not be sufficient for actually resolving the conflict.

    In many conflict situations, the existing alternatives or solutions may be inadequate or ineffective in addressing the underlying issues. This means that simply knowing how the problem owner perceives the problem may not lead to effective conflict resolution. Instead, there may be a need for new and previously unknown alternatives that go beyond the existing knowledge and understanding of the problem. Cognitive mapping may not be able to generate these new alternatives, as it primarily focuses on describing the existing cognitive structures of the problem owner(s).

    Moreover, cognitive mapping does not always provide information about the solution or suggest previously unknown actions about conflict drivers and how the problem owner configures the problem. In conflict transformation and management, it is crucial to not only understand the drivers of the conflict but also to generate creative and innovative solutions to address them. Cognitive mapping may fall short in providing guidance on how to actually resolve the conflict or generate new alternatives that were not previously known.

    Conflict transformation and management requires design-oriented methods beyond cognitive mapping. Design-oriented methods emphasize the need for creative problem-solving, generating new ideas, and designing interventions that can bring about positive change in the conflict dynamics. Cognitive mapping, being primarily descriptive in nature, may not fully support these design-oriented structures that are required for effective conflict transformation and management. In conclusion, while cognitive mapping can be a valuable tool in analyzing conflict drivers and understanding the problem owner's perspective, it may not provide innovative solutions or previously unknown alternatives for resolving the conflict.

\item Secondly, obtaining information about the source of a conflict is crucial for effective conflict transformation and management. Identifying the underlying source or root of a conflict allows for a more focused and solution-oriented approach in conflict analysis. By understanding the source of a conflict, it becomes possible to determine which conflict drivers are at play and develop comprehensive policies that address them effectively.

    However, defining the ``source of conflict'' can be challenging, as there is no universally agreed-upon definition. It can encompass diverse elements such as conflicting values, perceptions, and opposing tasks. Resolving the conflict source requires a deeper understanding of the intent and attitudes of stakeholders involved, and making sense of their motivations for seeking resolution. As shown in \cite{WatzlawickWeaklandFisch74}, changing a problem situation requires being able to modify how such a problem situation changes (changing how situations change) and for this purpose it is essential to identify the source and understand the dynamics driving a conflict and not just to describe it. By doing so, conflict resolution efforts can be more targeted and effective, addressing the root causes rather than just the symptoms of conflict. This underscores the importance of incorporating the concept of the conflict source into any cognitive mapping and other conflict analysis approaches for more comprehensive and solution-oriented conflict management strategies.

\item Thirdly, resolving conflicts requires finding a ``common ground'' among stakeholders, where similarities, shared concerns, and common interests can be identified. In principle, cognitive mapping, with its graphical representation of problems, serves a vital role in facilitating the discovery of connections and analysis of relations between concepts, enabling effective problem structuring. By visually mapping out the relationships between various elements of a problem, cognitive mapping helps to identify patterns, interdependencies, and causal relationships, providing valuable insights for understanding the complexity of the problem at hand. Cognitive mapping can aid in finding common ground by visually revealing areas of convergence and divergence among conflicting parties, providing a platform for meaningful discussions and identifying potential compromising points. The most common practice, however, consisting in constructing a single cognitive map for a group of concerned stakeholders does not seem to fit for severe conflict situations where even discussing together is questioned. The effectiveness of cognitive maps therefore depends on how they are used, for example, constructing a cognitive map for each stakeholder involved in the conflict allowing to understand how the conflict is perceived, understood and lived by each of the conflicting parties.

\end{enumerate}

Summarising, we argue that cognitive map structures for conflict transformation and management need to incorporate several essential elements: \\
1. These structures should go beyond the identification of known alternatives and also include the exploration of previously unknown alternatives, which can open up new possibilities for resolving conflicts. \\
2. Cognitive maps should enable a comprehensive understanding of the underlying issues at play, by facilitating the identification of the source or root causes of the conflict. \\
3. Cognitive maps should help identifying and highlighting common grounds or areas of agreement among conflicting parties, which can serve as a foundation for conflict resolution. \\
4. We consider necessary to start considering different cognitive maps for the different stakeholders involved in the conflict.

\subsection{Value trees}

Value-focused thinking (VFT) is a complex and multifaceted process that requires a formal structure for effective implementation (see \cite{Keeney92}). Keeney proposed several elements to identify values, including ethics, desired traits, characteristics of consequential outcomes, guidelines for action, priorities, value trade-offs, and attitudes toward risk. These elements serve as crucial indicators of the underlying values of a decision maker and require careful and deliberate consideration, often involving what Keeney referred to as ``hard thinking''(see \cite{Keeney92}, \cite{keeney1996valueejor}, \cite{keeney2007developing}, \cite{keeney1992value}). Although VFT methods are applied to reveal stakeholder values, their implementation steps do not align with the tree structure we are seeking (see \cite{gregory1994creating}). Furthermore, Keeney introduced the concept of ``means-end objective networks'' (\cite{keeney1994creativity}) as a method to identify objects in decision making, though it should be noted that this approach may not necessarily reveal the complete structure of the value tree, as identifying objectives and detecting values are distinct processes by definition. Objectives, as suggested in \cite{barber1990importance} are defined as actions that are required to achieve a specific goal whereas values, as defined by various scholars (see \cite{kluckhohn1951values}, \cite{rescher1969introduction}, \cite{znaniecki1940social}) can be understood as normative standards that exert influence over decisions, alternatives, or human beings. In essence, values play a pivotal role in shaping decision-making processes and outcomes, and are essential in understanding the ethical and normative dimensions of decision making, as emphasized in \cite{jacobi2007quantifying}.

Value-focused thinking, as proposed by Keeney, can lead to the generation of new and previously unnoticed alternatives as the decision maker begins to consider their values (see \cite{Keeney92}). The value-tree procedure is a method that allows individuals to systematically and hierarchically represent their values while engaging in problem-solving. However, some reservations exist in the literature regarding the representation of complex ideas using finite set procedures, as pointed out in \cite{dreyfus1992computers}.

Due to their hierarchical structure and single parent node property, value trees necessarily result in information loss and/or as noted in the literature attribute assimetry in various ways. For instance, a study examining the use of value trees in public policy analysis (see \cite{poyhonen2001behavioral}) argued that attributes in value trees can both positively or negatively affect the weight and rank of attributes. Similarly, \cite{poyhonen1998notes}, pointed out that normalized weighted averages may lead to false conclusions. In \cite{pitz1984content} the challenge of accurately describing values has been emphasised. It has equally be noted that top-down and down-top procedures may result in information losses.
As Keeney mentioned, conducting a semantic value analysis of a situation often requires complex thinking involving various elements, which may increase the likelihood of information loss due to changing attributes. Lakoff (see \cite{lakoff1982categories}) suggested using ``idealized cognitive models'' (ICMs) as an alternative to decompositional models for representing concepts, given the inherent difficulties in expressing complex meanings with decompositional models. However, \cite{pitz1984content} also noted that deriving idealized cognitive models and constructing the value tree of the decision-maker may not be feasible due to inconsistencies between ICMs.

Overall, the literature highlights both the benefits (see \cite{von1987value}) and limitations of value-focused thinking and the value tree procedure, including the potential for information loss and challenges in accurately representing complex values and decision-making processes.

Despite the structural reservations and lack of a universally agreed-upon formal method for constructing value trees, they offer a valuable way to visualize and analyze the values of a decision maker in a hierarchical manner. However, it should be noted that value trees may not fully capture the complexity of a problem. While they can be a useful tool for understanding and analyzing values, they may not provide a comprehensive description of a problem on their own. Other methods and techniques may need to be combined for a holistic and more comprehensive analysis, as value trees alone may not properly capture all the nuances and intricacies of a complex problem.

\subsection{Summary}

In summary, the literature on cognitive maps provides a useful structure for understanding the problem at hand, but falls short in providing solutions or strategies for handling the problem. As a descriptive model based on the problem owner's perceptions, cognitive maps do not necessarily offer (i) new alternatives, (ii) fail to identify the source of conflicts for comprehensive recommendations, and (iii) may not reveal common ground or shared concerns to mitigate conflict drivers. On the other hand, while value trees can be innovative for conflict transformation, they lack a formal structure and can result in informational asymmetry with top-down or bottom-up procedures, highlighting (i) the absence of a formal way to construct value tree structures, and (ii) how either top-down or bottom-up procedures create informational asymmetry. Despite their design-oriented structure, value trees may not fully capture the complexity of a problem in conflict resolution efforts.


\section{Our proposal}

Here we detail the transformation process from a cognitive map to a value cognitive map, then from a value cognitive map to an ends-means map, and then from an ends-mean map to a value tree.

\subsection{Notation}

In the following we will denote as a directed graph any set of the form $G=\{V,E\}$ where $V$ is a set of ``nodes'' represented as $V=\{x_1\cdots x_n\}$ and $E$ is a set of ``directed arcs'' represented as $E=\{x_ix_j: x_i,x_j\in V\}$. Clearly $E\subseteq V\times V$ and from a semantic point of view the arcs represent the pairs of elements in $V$ for which the binary relation $E$ holds.

\begin{definition}
Given a directed graph $G=\{V,E\}$ and two nodes $x_0,x_n\in V$ we define as a directed path from $x_0$ to $x_n$ a subgraph of $G$ such that $\exists x_1,\cdots,x_k: \exists x_0x_1,x_1x_2,$ $\cdots,x_kx_n$, where all $x_i$ are distinct.
\end{definition}

Given the directed graph $G=\{V,E\}$ and two nodes $x_0,x_n\in V$ if a directed path exists between $x_0$ and $x_n$ we denote it as $x_0-x_n$ or, abbreviated as $p_{0n}$ (just to differentiate it from the notation $x_0x_n$ which will represent a direct directed arc from $x_0$ to $x_n$). A directed graph is ``weakly connected'' iff for any given pair of nodes $x_i$ and $x_j$ there is a path $p_{ij}$ (not necessarily directed) connecting them. In other terms a directed graph is weakly connected if the underlying undirected graph is connected.


Given the directed graph $G=\{V,E\}$ and the arc $x_ix_j\in E$ we establish a positive real valued function $w:E\mapsto\mathds{R}^+$ and we name $w(x_ix_j)$ (or $w(ij)$ for abbreviation) as the ``length'' of the arc $x_ix_j$. Consider two nodes $x_0$ and $x_n$, since there may exist several paths between these nodes we denote a generic path between these two nodes as $p^l_{0n}$. Given a path $x_0-x_n$, we define the length of any path $p^l_{0n}$ as $w^l(0n)=\sum_{x_ix_j\in p^l_{0n}}w(ij)$. Given a path $x_0-x_n$ we denote $L(x_0-x_n)$ ($L(0n)$ for abbreviation) the cardinality of the set of nodes forming the path. In case all arcs have the same length $w(ij)=1$ then we have $w^l(0n)=L(0n)-1$. Independently of how the length of a path is computed we denote as $\hat{p}^l(0n)=\min_l(w^l(0n))$ the shortest path from node $x_0$ to node $x_n$.

\subsection{Cognitive Maps and Value Cognitive Maps}

\begin{definition}
  A cognitive map is a weakly connected directed graph $CM=\{N,R\}$ where $N$ is a set of ``concepts'' (or topics or issues ...) and $R=R^+\cup R^-$ is an ``influence'' relation composed by two binary relations such that:
  \noindent
  $R^+\subseteq N\times N$ and $R^-\subseteq N\times N$: \\
  $R^+(x,y)\;x,y\in N$ should read as ``$x$ has a positive influence on $y$'' \\
  $R^-(x,y)\;x,y\in N$ should read as ``$x$ has a negative influence on $y$''
\end{definition}

For the time being, the only constraint we impose is that both $R^+$ and $R^-$ are irreflexive relations, since intuitively ``$x$ has no influence upon $x$''.

We are now going to represent how to transform such a graph to a value tree. For this purpose we first need to create a new graph which we call a \textbf{Value Cognitive Map}. For this purpose we need to transform the set $N$ of concepts to a set $A$ of values. Practically we will associate to each concept used in the Cognitive Map a value (if this is possible). However, not all concepts can be translated into values, thus, $A\subseteq N$.

\vspace{5mm}

Although the semantics of the nodes has changed, we can still keep using the semantics of the relations $R^+$ and $R^-$. If $x$ and $y$ are elements of $A$ then: \\
 - $R^+(x,y)$ should read as ``value $x$ having a positive influence upon value $y$''; \\
 - $R^-(x,y)$ should read as ``value $x$ having a negative influence upon value $y$''. \\
When transiting from concepts to the realm of values, we introduce the prefix 'valuing' to signify the subjective nature intrinsic to the underlying values corresponding to their respective concepts. This approach aptly acknowledges the significance of negative narratives within the value cognitive map, underscoring the importance of analyzing both affirmative and adverse aspects, including the do's and don'ts, within the contextual framework. Although this linguistic choice may initially seem counterintuitive for certain concepts, it serves as a deliberate strategy to accentuate the value of comprehensive situational analysis and empower clients to expand their scope of self-expression.

\vspace{5mm}

Let's consider the graph $\{A,R\}$, $R=R^+\cup R^-$. In case this graph is weakly connected we can consider this to be a cognitive map and denote it as $VCM$ (Value Cognitive Map) in order to distinguish it from a regular Cognitive Map, since the set used here is a set of values.

\begin{definition}
  Consider a value cognitive map $VCM=\{A,R\}$ and a node $x_o\in A$ such that: $\forall x\in A\;\neg\exists R(x_o,x)$. We define $x_o$ as a fundamental node of the value cognitive map or a \underline{``fundamental value''}.
\end{definition}

\begin{proposition}
  Given a value cognitive map $VCM=\{A,R\}$ having a fundamental node $x_o$, there is always a directed path connecting any $x\in A$ to $x_o$.
\end{proposition}

\textbf{Proof}. Since the graph is weakly connected and by the definition of the fundamental node there is at least one arc such that $R(x,x_o)$ holds. Given any node of the graph, either it is a fundamental one and in this case we know there is path reaching it, or is a generic node. In this case, since the graph is weakly connected given the node $x_j$ there is always a node $x_i$ such that $R(x_i,x_j)$ holds. By induction on the number of nodes within the graph we show that there is always a directed path connecting the fundamental node $x_o$ to any node of the graph. \hfill$\blacksquare$

\vspace{5mm}

In other terms a fundamental value is influenced (directly or indirectly) by all values in the value cognitive map (since the graph is connected), but does not influence any other value.

\begin{definition}
  Given a value cognitive map $VCM=\{A,R\}$ and a generic node $x_j\in A$ we define as rank of $x_j$ (and we denote is as $r(x_j)$) the length of the shortest path from the node $x_j$ to the fundamental value $x_o$.
\end{definition}

Without loss of generality we assume that the length of any arc within the value cognitive map is equal to 1. In other terms the rank of any node in the value cognitive map is the number of nodes separating it from the fundamental value along the shortest path. Clearly, by definition, $r(x_o)=0$. From this point we only consider Value Cognitive Maps where exists a fundamental value.

\subsection{Ends Means Map}

\begin{definition}
  A Ends Mean Map is a weakly connected graph $EMM=\{B,\Pi\}$ such that $B$ is a set of ``values'' and $\Pi\subseteq B\times B$ is a binary relation to be read as follows: 
$\forall\chi,\psi\in B:\Pi(\chi,\psi)$ should be read as \textit{``value $\chi$ is an end to value $\psi$''}.
\end{definition}

The inverse should thus be understood as value $\psi$ being a mean to value $\chi$. In other terms the relation $\Pi$ is an ``ends-means'' relation upon the set of values $B$. \\
By the definition of ``ends-means'' the relation $\Pi$ needs to be irreflexive and acyclic (and therefore asymmetric). In the following we consider the problem: given a Value Cognitive Map how to transform it in an Ends Means Map.


We need to define how the relation $\Pi$ is constructed from the relation $R$. First of all we need to observe that semantically, if there is an influence relation (positive or negative) between $x_i$ and $x_j$ ($R(x_i,x_j)$: $x_i$ influences $x_j$) then the value represented by $x_j$ is an ``end'' to the value represented by $x_i$. In other terms, intuitively speaking, if $R(x_i,x_j)$ holds, then $\Pi(x_j,x_i)$ should also hold. However, the relation $R$ is composed by two relations (positive and negative influence, $R^+$ and $R^-$), while $\Pi$ is a unique relation. Further on, let's recall that a Value Cognitive Map is the graph $VCM=\{A,R\}$ and a Ends Means Map is the graph $EMM=\{B,\Pi\}$. We therefore need to establish a connection both between the sets $A$ and $B$ and the relations $R$ and $\Pi$.

Let's start introducing the set $\bar{A}=\{\neg x:x\in A\}$: the set of all negations of the elements in $A$. The set $B$ will be constructed from the union of $A$ and $\bar{A}$ respecting the weak connectivity property which the Ends Means Maps need to satisfy:
$B=\{x_i\in A\cup\hat{A}:\exists x_j\;\;\pi(x_i,x_j)\vee\pi(x_j,x_i)\}$.

\vspace{5mm}

\noindent
We give now the implications connecting the two relations. The reader should pay attention to the fact for these definitions $R^+\subseteq A\times A$, $R\subseteq A\times A$, $\Pi\subseteq A\cup\bar{A}\times A\cup\bar{A}$.

\begin{definition}~\\
 - $\forall x,y: R^+(x,y)\rightarrow\Pi(y,x)$ If $x$ has a positive influence upon $y$ then $y$ is an end to $x$ \\
 - $\forall x,y: R^+(x,y)\rightarrow\Pi(\neg y,\neg x)$ If $x$ has a positive influence upon $y$ then $\neg y$ is an end to $\neg x$\\
 - $\forall x,y: R^-(x,y)\rightarrow\Pi(y,\neg x)$ If $x$ has a negative influence upon $y$ then $y$ is an end to $\neg x$ \\
 - $\forall x,y: R^-(x,y)\rightarrow\Pi(\neg y,x)$ If $x$ has a negative influence upon $y$ then $\neg y$ is an end to $x$.
\end{definition}

\subsection{Algorithm}

We are now able to present the algorithm converting the value cognitive map $VCM=\{A,R\}$ to the ends means map $EMM=\{B,\Pi\}$ (see Algorithm \ref{alg01}).

\begin{algorithm}
\caption{Construction of a $EMM$}
\label{alg01}
 1. Import $A$ \\
 2. Import $R^+$ and $R^-$ \\
 3. Create $\bar{A}$ \\
 4. Label $x_o$ \\
 5. $\forall x\in A:\;\exists r^+(x,x_o)\rightarrow \pi(x_o,x)$ and eliminate $r^+(x,x_o)$ \\
 6. $\forall x\in A:\;\exists r^-(x,x_o)\rightarrow \pi(x_o,\neg x)$ and eliminate $r^-(x,x_o)$ \\
 7. Label all $x$ for which $\pi(x_o,x)$ \\
 8. Label all $\neg x$ for which $\pi(x_o,\neg x)$ \\
 9. $\forall x\mbox{ labelled }:\;\exists r^+(y,x)\rightarrow \pi(x,y)$ and eliminate $r^+(y,x)$ \\
 10. $\forall x\mbox{ labelled }:\;\exists r^-(y,x)\rightarrow \pi(x,\neg y)$ and eliminate $r^-(y,x)$ \\
 11. $\forall\neg x\mbox{ labelled and }x\mbox{ not labelled }:\;\exists r^+(y,x)\rightarrow \pi(\neg x,\neg y)$ and eliminate $r^+(y,x)$ \\
 12. $\forall\neg x\mbox{ labelled and }x\mbox{ not labelled }:\;\exists r^-(y,x)\rightarrow \pi(\neg x,y)$ and eliminate $r^-(y,x)$ \\
 13. Label all $y$ for which $\pi(x,y)$ or $\pi(\neg x,y)$ \\
 14. Label all $\neg y$ for which $\pi(x,\neg y)$ or $\pi(\neg x,\neg y)$ \\
 15. If no more $r^+(x,y)$ or $r^-(x,y)$ stop \\
 16. Eliminate all unlabelled nodes. \\
 17. For all cycles, if there is a unique longest path, eliminate the last arc. \\
 18. If there are more than one longest paths of the same length, submit to the client the choice of which are eliminate. \\
 19. Otherwise eliminate one arc of the cycle arbitrarily. \\
 20. End.
\end{algorithm}

\newpage
We begin by importing the concepts from set A, which constitute the nodes in the value cognitive map (1). Next, we establish all the existing relations within the value cognitive map (2). We include the negation of all concepts except the fundamental value (3). The process initiates with the fundamental value (4). Firstly, we consider all nodes that have a positive direct relation with the fundamental value. To induce an ends-means relationship, we reverse the direction of the arc from the fundamental value to the node, while eliminating the existing positive arc between the node and the fundamental value (5). Similarly, for nodes that have a direct negative relation with the fundamental value, we redirect the arc from the fundamental value to the negation of the node to ensure a positive relation, and we eliminate the existing negative relationship between the fundamental node and the node. It is important to note that with steps (5) and (6), we commence changing the direction of arrows and transforming signs.

In the subsequent steps, we label the nodes that have a direct positive (7) and negative (8) relationship with the fundamental node. Steps (9) and (10) involve examining the means of the labeled nodes. If a positive relation is identified, we modify the direction of the arc (9) (only changing the direction), whereas if a negative relation is found, we establish a relation between the negation of the mean (sign and direction simultaneously). It is worth noting that the process for the mean of the fundamental value follows the same procedure as outlined in steps (5) and (6).

Steps (11), (12), (13), and (14) require us to induce the same direction and sign transformation for the subsequent nodes. The key insight here is that we consistently change the direction of arrows and perform sign transformations based on altering the means while preserving the ends. The process halts when there are no more nodes to be processed (15). In the case of cycles, a unique situation arises as the process cannot conclude in the previous steps. We eliminate all nodes which we do not need any more (the unlabelled ones, Step (16)). Step (17) instructs us to eliminate relationships for the node that has the longest unique path, if such a node exists. If multiple longest paths exist, the client is consulted for elimination (18). Otherwise, one of the relations in the cycle is arbitrarily eliminated (19).

\noindent It is easy to prove the following propositions.

\begin{proposition}
  The algorithm \ref{alg01} converges in finite time.
\end{proposition}

\noindent\textbf{Proof}
  Straightforward. The algorithm contains no loops and each cycle (steps 5,6,9,10,11,12) is defined upon a finite set of nodes. The number of $r^+$ and $r^-$ arcs being finite it also takes a finite number of steps to construct the $\pi$ arcs. And this concludes the proof.
\hfill $\blacksquare$.

\begin{proposition}
  The graph constructed through algorithm \ref{alg01} has a unique fundamental node.
\end{proposition}

\noindent\textbf{Proof}
 Obvious. If the Cognitive Map has a unique fundamental node then the EMM will have a unique fundamental value.
\hfill $\blacksquare$.

\begin{proposition}
  The relation $\Pi$ of the graph constructed through algorithm \ref{alg01} is irreflexive and acyclic.
\end{proposition}

\noindent\textbf{Proof}
 Irreflexivity of $\Pi$ is a direct consequence of irreflexivity of $R$. There are no $R(x,x)$ arcs and therefore there cannot be construction of $\Pi(x,x)$ arcs. \\
 Acyclicity of $\Pi$ is obtained by construction. Cycles of $R$ arcs are eliminated substituting the $A$ set with the $B$ set ($B=A\cup \bar{A}$. Cycles not reduced by this step are further eliminated through steps 16, 17 and 18.
\hfill $\blacksquare$.

\begin{proposition}
  The graph constructed using algorithm \ref{alg01} is weakly connected.
\end{proposition}

\noindent\textbf{Proof}
 Since the graph has a unique fundamental value if for any node of the graph there is a path connecting it to such fundamental node then the graph is weakly connected. If a node is part of the graph (after applying the algorithm) it needs to be labelled. A node is labelled if there is an arc connecting it to a node already labelled or it is the fundamental value. Given any node either there is a path of labelled nodes connecting it to the fundamental node or it is the fundamental node. Thus, given any two nodes (labelled) both of them are connected to the fundamental value and thus there is an undirected path connecting them. The graph is weakly connected.
\hfill $\blacksquare$.

\begin{theorem}
  The graph constructed through algorithm \ref{alg01} up to Step 15 is unique with respect to the Value Cognitive Map from which it originates.
\end{theorem}

\noindent\textbf{Proof}
 Straightforward. Until Step 15 the algorithm is a sequence of deterministic steps and makes no choices. Thus, given any graph structure under form of a Value Cognitive Map there is a unique graph structure resulting from applying the algorithm.
\hfill $\blacksquare$.

\begin{corollary}
  The number of EMM constructed from a given VCM are finite and depend on how many cycles have to be reduced through algorithm \ref{alg01}.
\end{corollary}

\noindent\textbf{Proof} Straightforward.
\hfill $\blacksquare$.

\subsection{Value Trees}

Let us recall that a value tree is a graph structure representing ``ends-means'' relations satisfying the properties of a ``tree structure'':

\begin{definition}
  A directed graph $\langle G,V\rangle$ is a tree iff it is acyclic and for any two nodes there is a unique path connecting them.
\end{definition}

In case there is a single node connected to all nodes we call the tree structure an ``arborescence''. Under such a perspective a value tree as defined by Keeney is a ``value arborescence'': there is always a unique path form any value in the structure connecting it to the (unique) fundamental value (root) of the structure.

\vspace{5mm}

An EMM $\langle B,\Pi\rangle$ is an irreflexive, acyclic and weakly connected graph. Under such a perspective it can be seen as a structure with a unique ``root'' (the fundamental value), branching to ``layers'' of nodes, each layer being composed by nodes which are at the same distance from the root: the shortest path from the fundamental value to each node (recall that all arcs have the same length). Given the nodes of layer $k$ we call these predecessors of the nodes at layer $k+1$, these being successors of the nodes at layer $k$.

It is always possible that any given node seen as a ``mean'' (a successor) is related to multiple ``ends'' (predecessors). However, if this occurs, we do not have a tree structure since we do not satisfy the condition of having unique paths among any two nodes. We need to further elaborate the EMM in order to construct a value tree.

\vspace{5mm}

Let's consider a node $x_k$ (at layer $k$), having two predecessors: $x^1_{k-1}$ and $x^2_{k-1}$. We denote the fundamental node (the root) as $x_o$. Then there exist two paths from $x_o$ to $x_k$: \\
 - $(x_o-x_k)^1$: $\langle x_o\cdots x^1_{k-1}x_k\rangle$ and  \\
 - $(x_o-x_k)^2$: $\langle x_o\cdots x^2_{k-1}x_k\rangle$. \\
There are two cases. \\
1. Suppose $L(x_o-x_k)^1<L(x_o-x_k)^2$. We can conclude that what matters in terms of ends-means relation is the shortest path. This allows to eliminate the arc $x^2_{k-1}x_k$ since we consider it less relevant. Therefore, we can eliminate all ``predecessors'' of any $x_k$ which are on longest paths wrt to the fundamental node. \\
2. Suppose $L(x_o-x_k)^1=L(x_o-x_k)^2$. In order to explain better how we suggest handling this problem let us introduce two small examples.

\begin{example}
 Suppose $x_o$ is ``happiness'' having two means: $x^1_1$ (``health'') and $x^2_1$ (``pleasures''). Both these two nodes seen as ends have a mean (a successor) which is $x^1_2$ (``food''). We recall that each node of an EMM is a value representing preferences. We can now ask the user/client the following: \textbf{``are the preferences about food having an impact on health influenced by the preferences about food having an impact on pleasure?''}.

 \noindent If the reply is ``YES'' (i.e., I prefer junk food to bio food for pleasure, but I prefer bio food to junk food for health), then we can split the ``food'' node to two nodes: \\
  - $x^{11}_2$: food for health; \\
  - $x^{12}_2$: food for pleasure.
\end{example}

\begin{example}
 Suppose you go for dinner with your date. The success of the dinner ($x_o$), depends on three means: $x^1_1$ (food), $x^2_1$ (drinks) and $x^3_1$ (ambience). The value of food depends on whether it is $x^1_2$ (fish) or $x^2_2$ (meat), but also on what you drink (because food is differently appreciated if appropriately matched with drinks): $x^3_2$ (red wine) or $x^4_2$ (white wine). However, the same applies on the value of drinks (since the quality of the food has an impact on the appreciation of what you drink). The result is that all nodes $x^1_2,x^2_2,x^3_2,x^4_2$ share the same predecessors $x^1_1$ and $x^2_1$, all at the same distance from $x_o$.

 We can repeat the type of question of the previous example: \textbf{``are the preferences about fish or meat independent from your preferences about red or white wine?''} Most likely you will have a negative reply, the preferences between meat and fish being conditional on your preferences between red and white wine and viceversa. In this case the consequence will be merging the two predecessors $x^1_1$ and $x^2_1$ to a single node $x^{12}_1$ representing your preferences about combinations of food and drink. Once again we eliminated the multiple predecessors problem.
\end{example}

Let's summarise our presentation. Suppose node $x^j_k$ is a successor of multiple predecessors $x^{i}_{k-1}, x^{l}_{k-1}\cdots$. We can proceed as follows: \\
 - $\forall x^i_{k-1}\;$ find $i=\min_i L(x_o\cdots x^i_{k-1}x^j_k$. \\
 - Drop all arcs $x^l_{k-1}x^j_k$ such that $L(x_o\cdots x^l_{k-1}x^j_k) > L(x_o\cdots x^i_{k-1}x^j_k$). \\
 - If $x^i_{k-1}$ is unique, then stop, $x^j_k$ has a unique predecessor. \\
 - If there are more than one $x^i_{k-1}$ ($i,i',i''\cdots$) then: \\
 $\bullet$ If preferences in $x^j_k$ as far as $x^i_{k-1}$ are concerned are independent from the preferences in $x^j_k$ as far as $x^{i'}_{k-1}$ are concerned, then split $x^j_k$ in nodes $x^{ji}_k$ and $x^{ji'}_k$. \\
 $\bullet$ If preferences in $x^j_k$ as far as $x^i_{k-1}$ are concerned are conditions to the preferences in $x^j_k$ as far as $x^{i'}_{k-1}$ are concerned (and/or viceversa), then merge $x^i_{k-1}$ and $x^{i'}_{k-1}$ in $x^{ii'}_{k-1}$.

\vspace{5mm}

This procedure will lead (in finite time) in eliminating all multiple predecessors for each successor within a $EMM=\{B,\Pi\}$ . The result will be a new graph $VT=\{V,\Pi\}$ where the set $V$ is obtained from set $B$ by merging or splitting nodes and the relation $\Pi$ is obtained eliminating or adding arcs as specified.

%
%
%

\section{Case Study}

The protracted and intricate armed conflict between Turkey and the Kurdistan Workers' Party (PKK) has a complex and extensive historical background stemming from Kurdish revolts (see \cite{yadirgi2017political}). Since its inception as a nation-state, Turkey has struggled with the Kurdish question, which presupposes that all citizens are Turkish (see \cite{barics2014turkluk}), rendering it arduous for minority groups, including the Kurds, to maintain their distinct identity. Consequently, Turkey has adopted assimilative/oppressing policies that have suppressed Kurdish culture, language, and identity, contributing to the escalation of the Kurdish question into an armed conflict in 1984 (see \cite{bozarslan2008kurds}).

The origins of the conflict can be attributed to historical and cultural tensions between the Turkish state and the Kurdish people, as well as geopolitical factors such as regional power struggles and external influences (see \cite{ccandar2020turkey}). Despite numerous attempts to resolve the conflict, including two unsuccessful ceasefires in 2015, it continues to be a pressing issue that poses significant challenges for Turkey and the broader Middle East region, marked by numerous human rights violations. For instance, in May 2023, many Kurdish politicians and human rights defenders were incarcerated.

Two stakeholders, whose identities are kept anonymous, were selected for this study. One is an academician who specializes in Kurdish issues and represents the Kurdish side, while the other is a high-ranking politician from the main opposition party, who represents the Turkish side. In order to identify the core motivation of the conflict, a series of comparative questions were asked, and both stakeholders confirmed the specified fundamental node of the conflict. It should be noted that the knowledge of only two stakeholders cannot be decisive in a conflict as wide-ranging as the Kurdish-Turkish one. This study aims to test the effectiveness of the proposed model and not to suggest a real solution to this, far more complicated, conflict.

The connections and information provided by the stakeholders in the interviews are consistent with other studies on the Kurdish-Turkish conflict found in the literature (see \cite{bozarslan2008kurds}, \cite{ccandar2020turkey}). The primary objectives of this study is to analyze the problem in greater depth using cognitive maps, identify the connections between values using value cognitive maps, and examine the value-based structure of the problem while exploring the possibility of discovering new alternatives with value trees. In summary, the study utilized cognitive mapping to analyze the Kurdish-Turkish conflict thoroughly, and the first step was to transform the cognitive map into a value tree. The output of the value tree not only enhances problem structuring but also enables the exploration of new alternatives. Cognitive maps allow for the analysis of problems, concerns, solutions, and past actions, while value trees reveal shared or conflicting values. Lastly, the procedure allows for the generation of unknown alternatives by thinking in terms of values.

\subsection{Kurdish Perspective}

\subsubsection{The story}

\begin{center}
\begin{figure}[H]
 \scalebox{0.95}{\includegraphics[width=13cm]{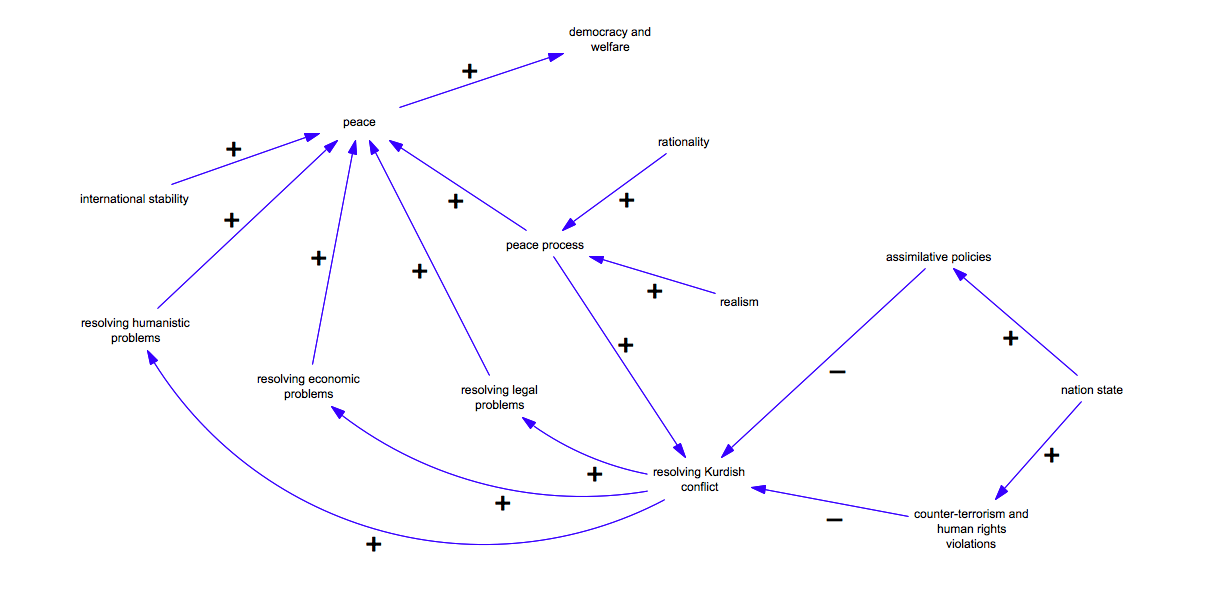}}
 \caption{The Kurdish cognitive map}\label{kurdcm}
\end{figure}
\end{center}

The cognitive map (figure \ref{kurdcm}) was constructed based on an interview conducted with an expert academic who specializes in the Turkish-Kurdish conflict. The perspective of the Kurdish stakeholder can be summarized as follows: the fundamental node is ``Democracy and welfare'', and all nodes within the cognitive map have outgoing arcs leading to this fundamental node. The fundamental node itself is unidimensional and primarily focuses on the concept of``"peace''. According to the stakeholder, attaining peace necessitates a ``peace process'' and ``international stability'' and resolving humanitarian, economic, and legal problems. To ensure an effective peace process, the stakeholder emphasizes the importance of two values: ``rationality'' and ``realism''. The Kurdish conflict gives rise to various issues, including ``economic'', ``humanitarian'' and ``legal problems''. Consequently, resolving the Kurdish conflict would have a positive impact on addressing economic, legal, and humanitarian concerns. The notion of the Turkish nation-state encompasses ``assimilative policies'', and its approach to ``counter-terrorism'' has resulted in ``human rights violations''. From the stakeholder's perspective, these two dimensions of the nation-state are viewed as underlying factors contributing to the Kurdish conflict.

\subsubsection{Transformation Process: Cognitive map to Value cognitive map}

We are required to transform the set N, consisting of concepts, into a set A comprising values. In practical implementation, it is our endeavor to assign a value to each concept employed in the Cognitive Map, whenever it proves feasible. Nevertheless, it is imperative to acknowledge that not all concepts can be effectively transposed into values. In the context of the Kurdish Cognitive Map, all concepts can indeed be translated into corresponding values. Moreover, realism and rationality were already values in the form expressed by the stakeholder. Take note that the concept 'peace' is not a value here; instead, it is used as a term to describe the desire to achieve peace, i.e. an objective.

\begin{center}
\begin{figure}[H]

    \scalebox{0.30}{\includegraphics{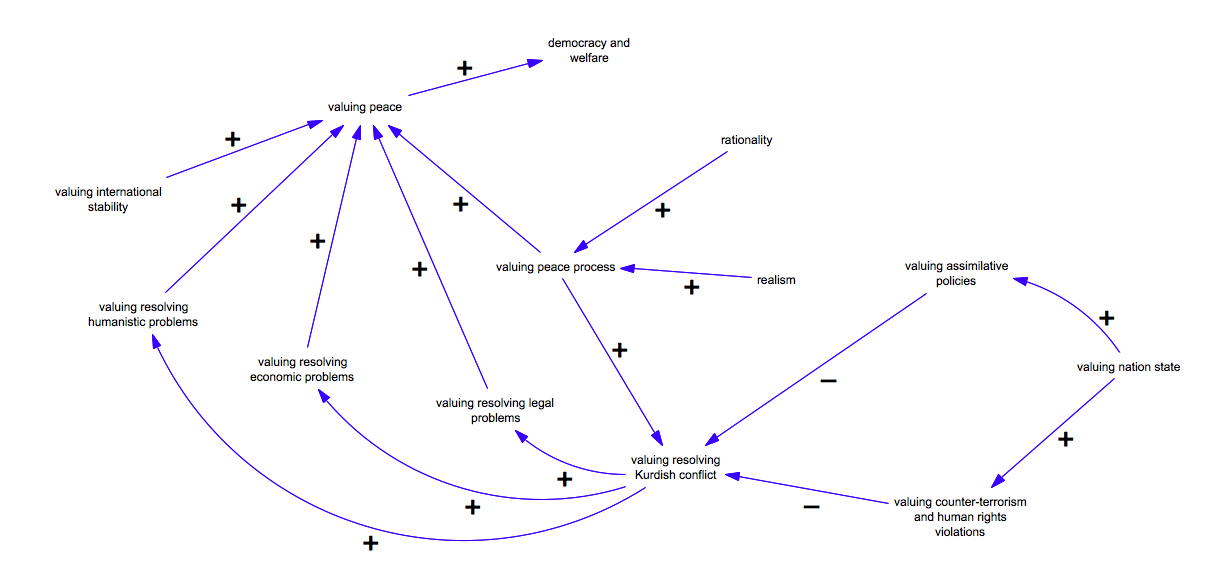}}
    \caption{The Kurdish Value Cognitive Map}\label{kurdvcm}
\end{figure}
\end{center}

Within the Kurdish value cognitive map (figure \ref{kurdvcm}), the fundamental value is ``democracy and welfare'', whereby ``valuing peace'' exerts a positive influence on ``valuing democracy and welfare''. Conversely, the presence of negative arcs in the map signifies that ``valuing nation-state'' will have a positive effect on counter-terrorism and human rights violations, while ``valuing assimilative policies'' leads to a negative influence on resolving the Kurdish conflict. It is worth noting that the use of the ``valuing'' prefix allows for the exploration of both positive and negative narratives. For instance, ``valuing assimilative policies'' can be interpreted as a desire to maintain such policies, and by examining the influence diagram, we can gain insights into the potential consequences of such values.

\subsubsection{Transformation process: Value cognitive map to Ends-Mean Map}

The next step involves elucidating the process of constructing the relation $\Pi$ from the relation $R$. It is crucial to highlight that when an influence relation (positive or negative) exists between $x_i$ and $x_j$ ($R(x_i,x_j)$: $x_i$ influences $x_j$) in terms of semantics, the value represented by $x_j$ assumes the role of an ``end'' to the value represented by $x_i$. However, it is important to note that while the relation $R$ encompasses two distinct relations, namely the positive influence ($R^+$) and the negative influence ($R^-$), the relation $\Pi$ solely comprises positive influences. Consequently, our algorithm commences by altering the direction of arrows, facilitating the induction of ends-means relationships, and subsequently undertaking the necessary sign transformations from negative to positive.

\begin{center}
\begin{figure}[H]
    \scalebox{0.30}{\includegraphics{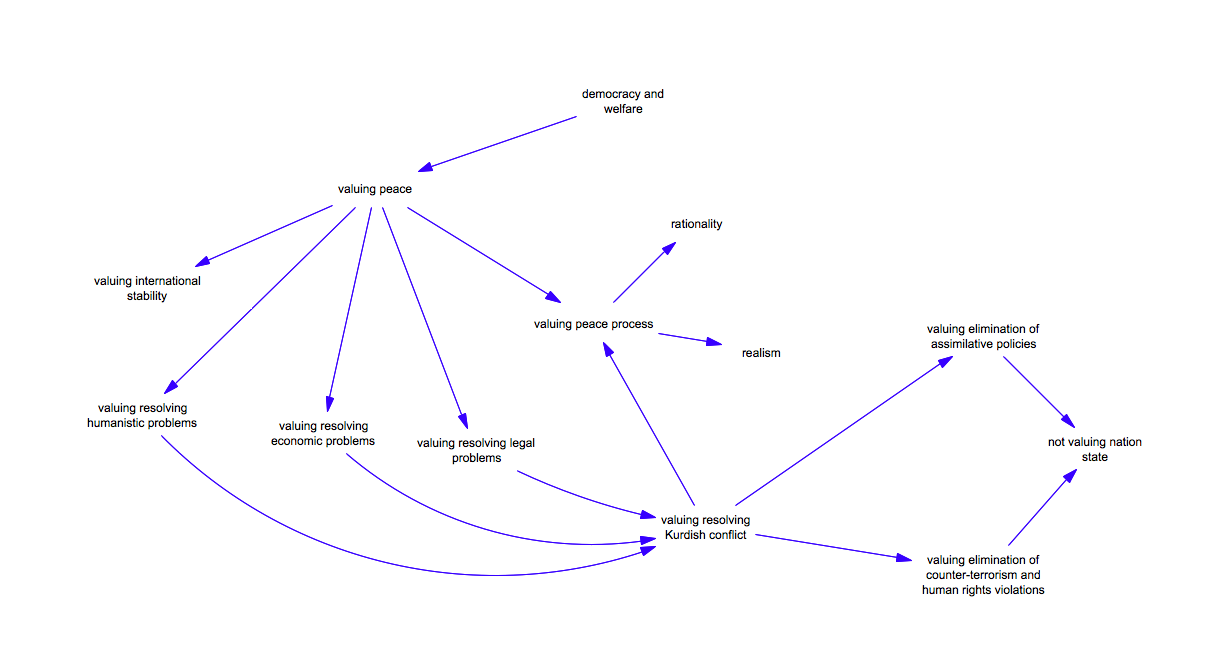}}
    \caption{The Kurdish Ends-Means Map}\label{kurdemm}
\end{figure}
\end{center}

We start duplicating all nodes to ensure that their negations are available for sustaining sign transformation. In our case, we observe two negative relationships: one between ``valuing assimilative policies'' and ``valuing resolving Kurdish conflict'', and the other between ``valuing counter-terrorism and human rights violations'' and ``valuing resolving Kurdish conflict''. According to the EMM (Ends-Means-Mapping) framework, these negative relationships need to be eliminated. Since ``valuing resolving Kurdish conflict'' serves as the ends for both relations, we choose to negate ``valuing assimilative policies`` and ``valuing counter-terrorism and human rights violations'', as they represent the means in question (see figure \ref{kurdemm}).

It is crucial to highlight the advantages of duplicating nodes along with their negations. By utilizing ``valuing elimination of assimilative policies'' instead of ``valuing assimilative policies'', we maintain the positive relationship. Similarly, by using ``valuing elimination of counter-terrorism and human rights violations'', we achieve the desired sign transformation. However, it should be noted that when we choose the negations for these two nodes, we lose the positive relationship between their non-negated versions and ``valuing nation state''. To address this issue, we also negate the concept of valuing the nation state by using ``not valuing nation state''. It is important to recognize that while changing the narrative can alter the signs, our algorithms always prioritize the ends over the means.

\subsubsection{Transformation process: Ends-Mean Map to Value Tree}

\begin{center}
\begin{figure}[H]
    \scalebox{0.3}{\includegraphics{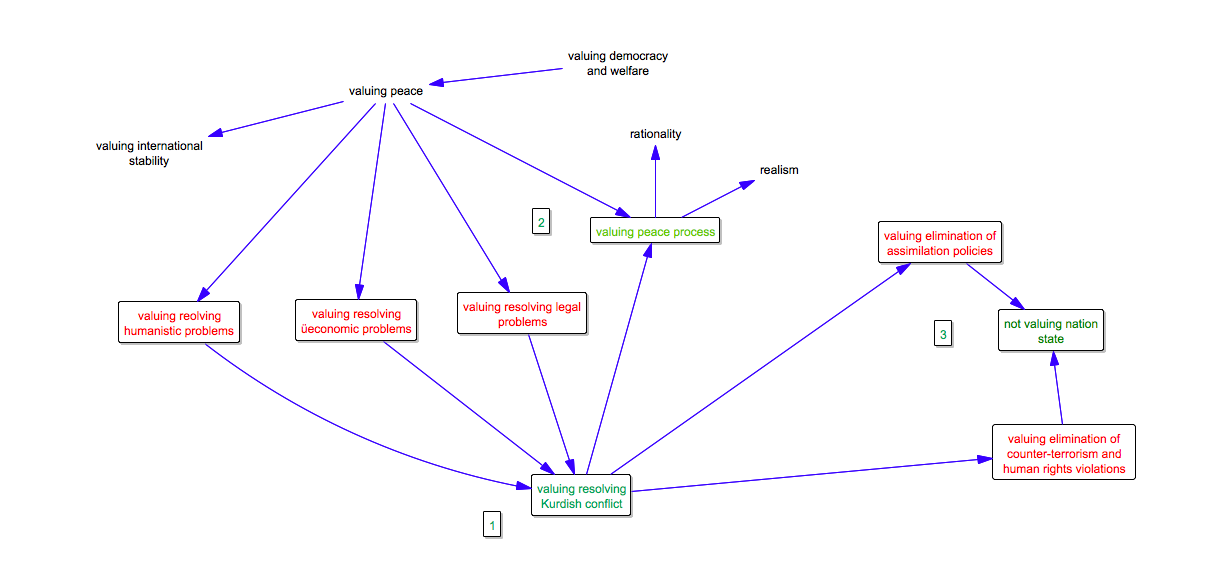}}
    \caption{Multiple predecessors problem}\label{kurdemmvt}
\end{figure}
\end{center}

The next step consists in establishing the value tree of the Kurdish stakeholder. The Kurdish EMM does not respect the tree structure because
we have three nodes (highlighted in green) that have multiple predecessors (see figure \ref{kurdemmvt}). According to our process, the first step is to examine these predecessors (highlighted in red) in terms of their shortest path to the fundamental value.

For Node 1, which represents valuing the Kurdish conflict, there are three predecessors: ``valuing resolving legal problems'', ``valuing resolving economic problems'', and ``valuing resolving humanitarian problems''. These three nodes share the same length of the shortest path to the fundamental value. As a result, we cannot determine their order based on that information. The next step in our procedure is to assess whether these predecessors have independent preferences. Since all these predecessors are created by the Kurdish conflict in the cognitive map, their preferences over means are not independent. Therefore, instead of splitting the means, we merge the ends and create a new node called ``valuing resolving general problems''.

For Node 2, ``valuing peace process'', there are two predecessors: ``valuing peace'' and ``valuing resolving Kurdish conflict''. We determine the shortest path based on the fundamental value. In this case, the shortest path is from ``valuing peace''. Therefore, we eliminate the arc between ``valuing resolving Kurdish conflict`` and ``valuing peace process''.

Similarly, for Node 3, which represents ``valuing the not nation-state'', there are two predecessors: ``valuing elimination of assimilative policies'' and ``valuing elimination of counter-terrorism and human rights violations''. Like the predecessors of Node 1, these predecessors also have the same shortest path to the fundamental value, making it difficult to determine their order based on distance alone. Once again, we assess whether there are independent preferences among the predecessors. However, we find that their preferences are dependent. As a result, we merge these two ends and create a new node labeled as ``valuing elimination of oppressing policies''.

Finally we end up with the Kurdish value tree (figure \ref{kurdvt}):

\begin{center}
\begin{figure}[H]
    \scalebox{0.45}{\includegraphics{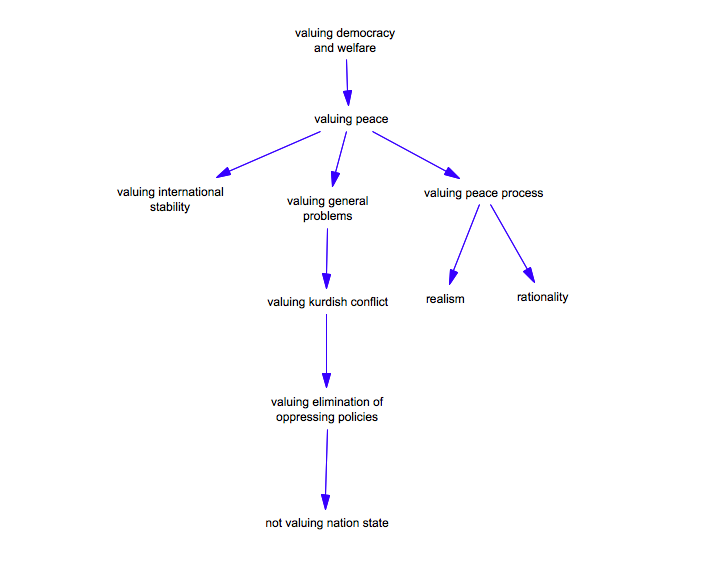}}
    \caption{The Kurdish value tree}\label{kurdvt}
\end{figure}
\end{center}

\subsection{Turkish Perspective}

\begin{center}
\begin{figure}[H]
    \scalebox{0.30}{\includegraphics{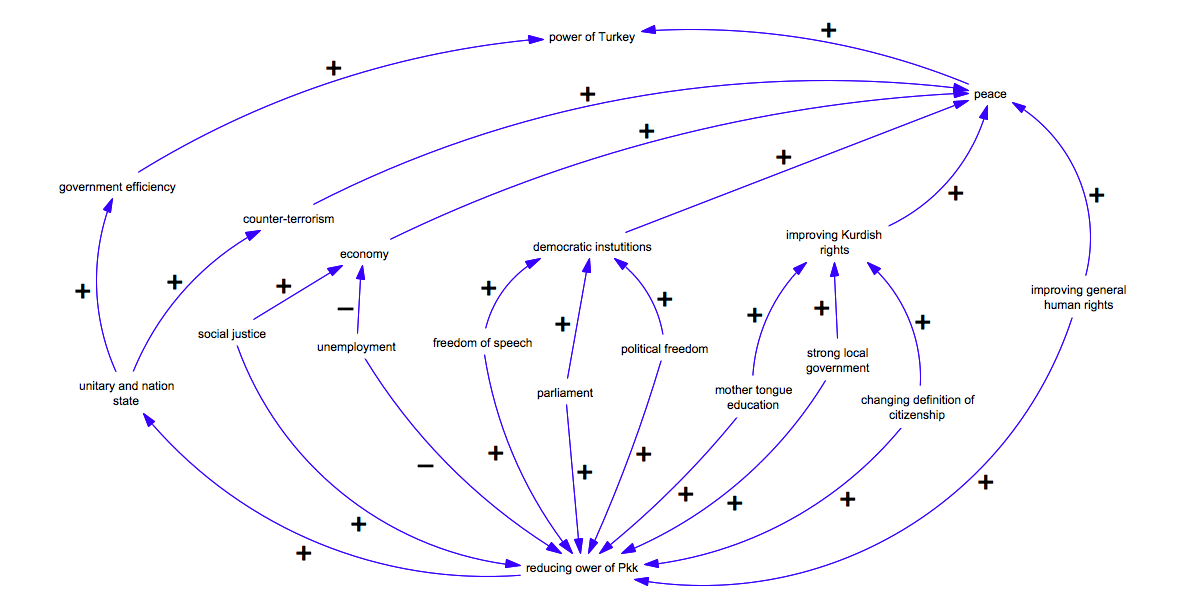}}
    \caption{The Turkish Cognitive Map}\label{turkcm}
\end{figure}
\end{center}

\noindent The interviewed stakeholder provided insights that helped create a cognitive map of the Turkish stakeholder's perspective (see figure \ref{turkcm}). The fundamental value for the stakeholder is ``The Power of Turkey'', which consists of two main legs: ``Peace'' and ``Government Efficiency''. While the stakeholder explicitly highlighted peace as a vital value and concept, it primarily serves the well-being of Turkey as a whole. Therefore, to achieve peace, Turkey must improve several aspects, including ``Kurdish rights', ``general human rights'', ``democratic institutions'', and ``economy'' while continuing counter-terrorism actions.

The ``Government Efficiency'', which is essential for the ``Power of Turkey'', is sustained by the ``Unitary and Nation-State'' structure. This structure has a positive effect on counter-terrorism, which is vital for peace. However, the PKK has a substantial adverse effect on the ``Unitary and Nation-States'' node. While every sub-node for ``peace'' harms the ``PKK'', except for the ``counter-terrorism'' node, improvements in ``democratic institutions'' require an improvement in ``freedom of speech'', ``parliament'', and ``political freedom`''. All three nodes harm the ``Power of PKK''. Therefore, the improvements in the ``economy'', ``democratic institutions'', ``general human rights'', and ``Kurdish rights'' will not only reduce the ``Power of PKK'', which harms the ``Unitary State'', but they will also enhance peace as a whole subnode.

In summary, the ``Power of Turkey'' has two primary benchmarks: ``Peace'' and ``Government Efficiency'', while the ``Power of PKK'' can harm the ``Unitary State'' before ``Government Efficiency'' and ``Peace'', improving the nodes of peace will not only enhance peace itself, but it will also decrease the power of PKK and prevent harm to the Government Efficiency (see figure \ref{turkcm}).

\subsubsection{Transformation Process: Cognitive map to Value cognitive map}

\begin{center}
\begin{figure}[H]
    \scalebox{0.30}{\includegraphics{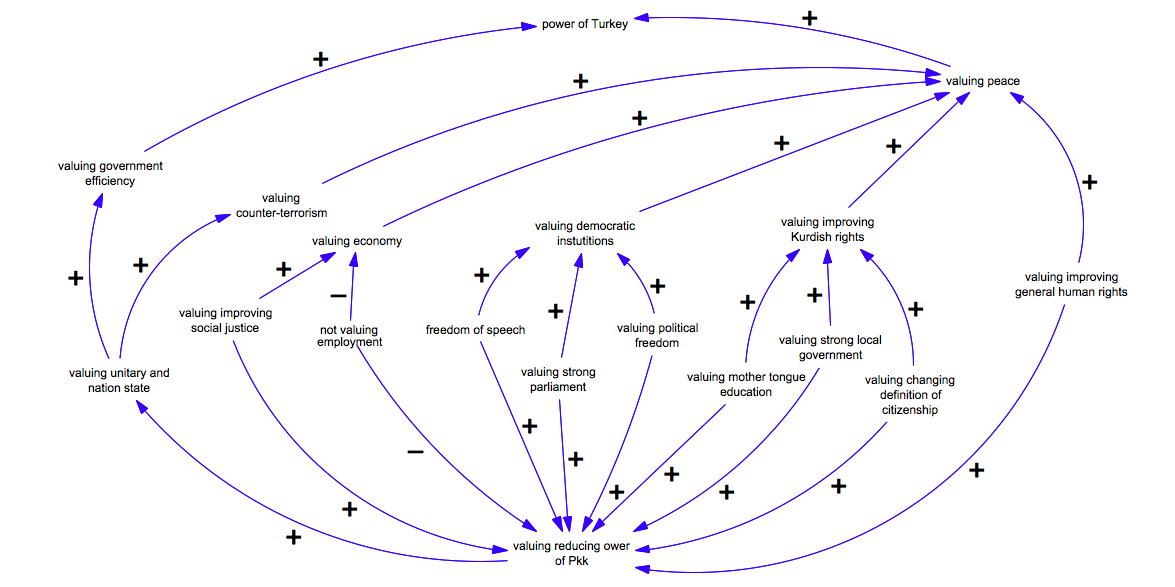}}
    \caption{The Turkish Value Cognitive Map}\label{turkvcm}
\end{figure}
\end{center}

We need to undertake the task of converting the set N, which encompasses various concepts, into a set A consisting of values. In the Turkish cognitive map, all concepts can be represented as values. Furthermore, we have already incorporated three values into the map: social justice, freedom of speech, and general human rights. It is worth noting that, similar to the Kurdish cognitive map, the concept of peace in the Turkish map is utilized to describe the attainment of peace rather than representing a value in itself (see figure \ref{turkvcm}).

In the Turkish value cognitive map, it is evident that valuing peace and valuing the ``Government Efficiency'' contribute to enhancing the power of Turkey, which serves as the fundamental value. Conversely, the concept of ``not valuing employment'' (replacing unemployment as the negation of valuing employment) is associated with a negative impact on ``valuing reducing power of PKK''.

\subsubsection{Transformation process: Value cognitive map to Ends-Mean Map}

\begin{center}
\begin{figure}[H]
    \scalebox{0.30}{\includegraphics{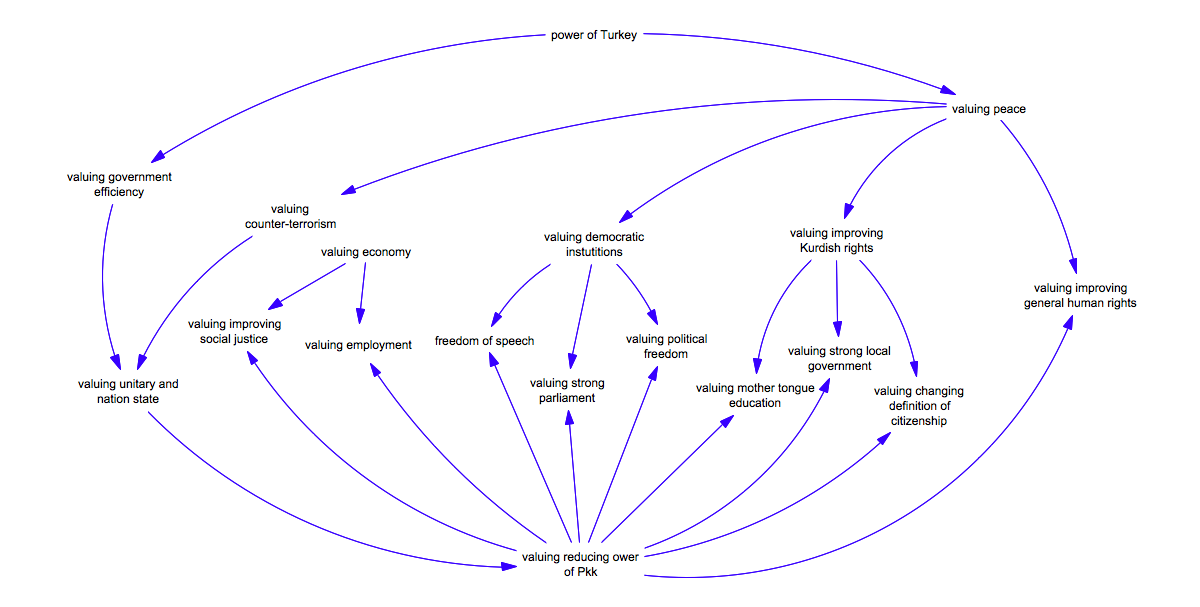}}
    \caption{The Turkish Ends-Means Map}\label{turkemm}
\end{figure}
\end{center}

Similar to what we carried out for the Kurdish side, we proceed with changing the direction of arrows to maintain the ends-means relationship and facilitate sign transformation as required for the ends-means relationship. Our algorithms dictate the duplication of nodes and the establishment of all positive relationships from ends to means by utilizing the negation of means at each stage. As there were no cycles present, the transformation process proceeded straightforward (see figure \ref{turkemm}).

In the case of ``valuing reducing power of PKK'', which exhibited a positive influence on all concepts except for ``not valuing employment'', we employed the negation of ``not valuing employment'', represented as ``valuing employment''. This choice of narrative also resulted in a change of the negative sign between ``not valuing employment'' and ``valuing economy''. By duplicating the node and considering the appropriate narrative, we ensure that our process aligns with our underlying intuition, which prioritizes preserving ends over means at each stage.

\subsubsection{Transformation process: Ends-Mean Map to Value Tree}

We have identified ten nodes (highlighted in green in figure \ref{kurdemmvt}) that had multiple predecessors. Following our algorithm, the initial step entails examining these predecessors in terms of their shortest path to the fundamental value.

\begin{center}
\begin{figure}[H]
    \scalebox{0.30}{\includegraphics{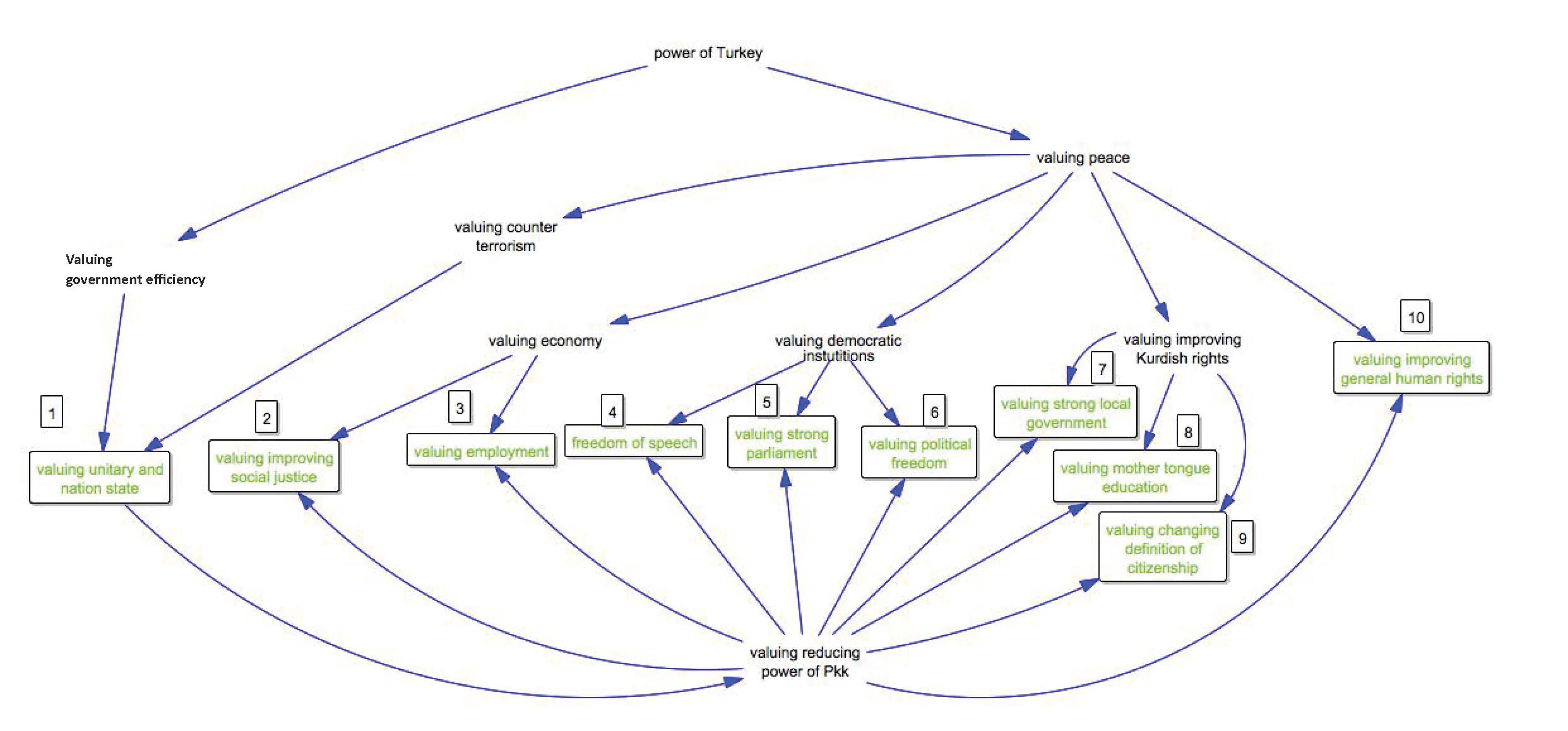}}
    \caption{Multiple Predecessors Problem}\label{turkemmvt}
\end{figure}
\end{center}

In the first case of multiple predecessors (Case 1), we examine the shortest path between ``valuing Government Efficiency'' and ``valuing counterterrorism''. Since the shortest path belongs to ``valuing Government Efficiency'', we eliminate the arc between ``valuing counterterrorism'' and ``valuing unitary and nation state''.

In all other cases of multiple predecessor problems, one of the predecessors is ``not valuing power of PKK''. However, it never possesses the shortest path compared to the other predecessors. Therefore, we eliminate the arcs between ``valuing reducing power of PKK'' and the following nodes: ``social justice'' (Case 2), ``valuing employment'' (Case 3), ``freedom of speech'' (Case 4), ``valuing strong parliament'' (Case 5), ``valuing political freedom'' (Case 6), ``valuing mother tongue education'' (Case 7), ``valuing strong local government'' (Case 8), ``valuing definition of citizenship'' (Case 9), and ``valuing general human rights'' (Case 10). We end up with the Turkish value tree (see figure \ref{turkvt}):

\begin{center}
\begin{figure}[H]
    \scalebox{0.25}{\includegraphics{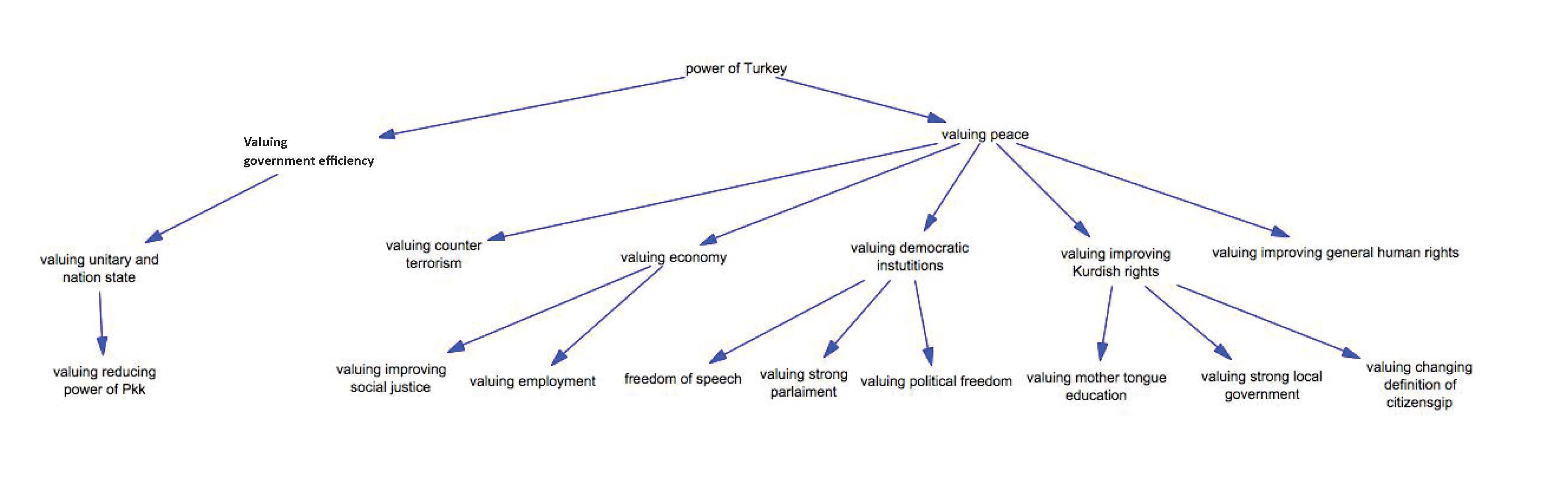}}
    \caption{The Turkish Value Tree}\label{turkvt}
\end{figure}
\end{center}

\subsection{Discussion}

Supposing the two cognitive maps (represented in figures \ref{kurdcm} and \ref{turkcm}) belong to two stakeholders with the relevant political power to engage a negotiation process, what could be the suggestion of an external advisor who has access to this knowledge? As they stand it seems there is no common ground for even starting a discussion between the two conflicting stakeholders. The reason is that a simple descriptive representation of how the problem situation is perceived by the stakeholders does not reveal which are the drivers of the two parts. As the situation stands presently the two parts only recognize the threats the other part presents for their fundamental concerns (for the Kurdish the repressive Turkish policy and for the Turkish the Kurdish terrorism).

The value structure instead reveals that there is a common ground for starting discussion among these two stakeholders (see figures \ref{kurdvt} and \ref{turkvt}). The exact point of possible convergence is reducing the repressive policies (for the Kurdish) and improving the democratic institutions (for the Turkish). Although at this time these two topics might appear relative distant they represent a possible point of convergence upon which to try to build a dialogue searching actions for reducing and transforming the conflict.

The value tree also reveals which is the potentially critical point for finding a long term convergence changing the conflict in a sustainable (long term) way: the definition of citizenship for the Turkish democracy. If the notion of citizen is based on a pure nationalistic narrative (as it has been for both sides for the whole duration of the conflict) then there are little chances any discussion (even in good will) could deliver a sustainable solution on the long run. If the two parts instead are ready to explore the notion of Turkish citizen on the basis of shareable human and democratic rights then there are grounds for a long term solution.

The value tree generated through the process offers opportunities for innovative policy alternatives. For instance, when considering the Turkish side's value of ``valuing mother tongue education'', we can explore novel solutions that enhance mother tongue education for Kurds without compromising the integrity of the nation-state. Noting values related to ``valuing strong local government'' and ``valuing nation state'', an option such as adopting the ``European Charter of Local Self-Government''(see \cite{df1c7742-7905-3519-823b-391e4bdbae11}) could emerge as a viable alternative. This approach encourages the exploration of creative and balanced solutions to address complex conflicts.

\section{Conclusion}

In conclusion, this paper presents a significant contribution to the field of conflict transformation and management through the integration of cognitive mapping and value trees. The combination of these two well-known problem structuring methods fills the gap between descriptive and design-oriented approaches, providing decision-makers with a more holistic understanding of conflicts and innovative solutions. Additionally, our transformation process enables the construction of value trees from cognitive maps and establishes a formal method for this purpose, introducing a novel approach in itself.

The transformation process from cognitive maps to value trees allows for a deeper analysis of the relationships between values, offering a structured representation that aids in policy design and decision-making. By extracting a value tree from the cognitive map, we leverage the rich qualitative information captured in cognitive maps and translate it into a more structured and formal representation usable for prescriptive purposes. This formal method of constructing value trees is a noteworthy contribution to the field, as it facilitates a systematic approach to explore creative and previously unknown solutions.

The case study on the Kurdish-Turkish conflict demonstrates the effectiveness and replicability of our method in a real-world conflict scenario. By identifying common ground and potential areas of compromise, our approach fosters collaborative thinking and constructive issue resolution. This not only makes it relevant for conflicting situations but also extends its applicability to decision-aiding and policy design in other complex contexts.

Looking ahead, future research can explore further applications of this method in diverse conflict contexts, extending its practical value and relevance. By providing decision-makers with a comprehensive tool that combines problem structuring and creative solution generation, our approach contributes to advancing conflict studies and facilitating sustainable conflict resolution strategies.

\bibliographystyle{plain}
\bibliography{berkay}

\end{document}